# Characterizing Design Patterns of EHR-Driven Phenotype Extraction Algorithms

Yizhen Zhong, Luke Rasmussen, Yu Deng, Jennifer Pacheco, Maureen Smith, Justin Starren
Feinberg School of Medicine
Northwestern University
Chicago, IL

Wei-Qi Wei, Peter Speltz, Joshua Denny
Dept. of Biomedical Informatics
Vanderbilt University
Nashville, TN

Nephi Walton
Genomic Medicine Institute
Geisinger
Danville, PA

George Hripcsak
Dept. of Biomedical Informatics
Columbia University
New York, NY

Christopher G Chute
Schools of Medicine, Public Health, Nursing
Johns Hopkins University
Baltimore, MD

Yuan Luo* (Corresponding)
Dept. of Preventive Medicine
Northwestern University
Chicago, IL

*Abstract:* **The automatic development of phenotype algorithms from Electronic Health Record data with machine learning (ML) techniques is of great interest given the current practice is very time-consuming and resource intensive. The extraction of design patterns from phenotype algorithms is essential to understand their rationale and standard, with great potential to automate the development process. In this pilot study, we perform network visualization on the design patterns and their associations with phenotypes and sites. We classify design patterns using the fragments from previously annotated phenotype algorithms as the ground truth. The classification performance is used as a proxy for coherence at the attribution level. The bag-of-words representation with knowledge-based features generated a good performance in the classification task (0.79 macro-f1 scores). Good classification accuracy with simple features demonstrated the attribution coherence and the feasibility of automatic identification of design patterns. Our results point to both the feasibility and challenges of automatic identification of phenotyping design patterns, which would power the automatic development of phenotype algorithms.**

## I. Introduction

Phenotype algorithms are designed to enable robust selection of patients that meet certain research interests from the Electronic Health Record (EHR) system [1, 2]. The electronic Medical Records and Genomics (eMERGE) network [3] is an NHGRI-sponsored initiative to further the development and implementation of EHR-derived phenotypes across multiple institutions. The phenotype algorithms developed, validated and implemented through multi-site collaboration are primarily recorded as text documents [4] and are re-implemented in a computable format at each site. The Phenotype Knowledge Base (PheKB) [5] is a primary source to store EHR derived phenotyping algorithms in eMERGE.

The development of phenotype algorithms is a routine practice carried out by biomedical informaticians by mining the EHRs. The development process is complicated not only by the nuances of how data is collected within the EHR but also by the heterogeneity across EHR vendors, implementations and individual use. Awareness of these differences and accounting for them up front in the phenotype development process may increase the efficiency of the development process, and the accuracy and portability of the resulting phenotypes. To facilitate efficient and reproducible phenotype algorithm development, we previously proposed "phenotype design patterns" [6]. A phenotype design pattern is intended to address commonly seen inputs, logic and constraints in the phenotyping algorithms, and provide guidance to a solution. Design pattern annotations provide a deep understanding of organizations and semantic and syntactic structures of phenotyping algorithms and are helpful for rationalizing the philosophy, standardizing and even automating the development process of phenotyping algorithms.

The initial set of phenotype design patterns was developed by an iterative review of existing phenotype algorithms and leveraged opinions from multiple experts. Scaling such tasks to large sets of phenotyping algorithms is laborious, time-consuming, and typically requires experts with domain knowledge. There is a need to automate the design pattern annotation process. Here, we aim to quantitatively characterize the coherence of the design patterns to assess both the practicability and challenluoges of design pattern attribution. Built on the work of Rasmussen et al., we aim to 1) visualize the design patterns and their associations with phenotypes and sites; 2) classify design patterns as a proxy for assessing their attribution coherence.

## II. Methods

### A. Network Visualization

Rasmussen et al. annotated the narrative description of each phenotype algorithm downloaded from Phenotype Knowledge Base (PheKB.org) [3], extracted ~250 sentences that described aspects of the data elements and/or logic used in the algorithm, and summarized design patterns, with information of phenotype and developing site available. Using that data set, in this work we visualized the unique combinations of design pattern, phenotype and site in three networks with Google Fusion Table [7]: a) phenotype and site, b) phenotype and design pattern, and c) site and design pattern. The size of the node is proportional to the number of occurrences in the dataset and the width of the edge corresponds to the number of connections between the associated nodes.

* Corresponding yuan.luo@northwestern.edu. This study was supported in part by NIH grant R21LM012618 and 5U01HG008673.

Table 1. Phenotyping design patterns and descriptions.

| Pattern | Description | Training | Testing |
|---|---|---|---|
| Where Did It Happen | Knowing if something was inpatient or outpatient is important. Kind of like Transient Conditions, if the patient is in the hospital, we exclude data in many cases. | 14 | 5 |
| Credentials of the Actor | If you need a physician to make a diagnosis, make sure a physician entered it. Likewise, if you need a specialist to make the diagnosis, ensure that is the data you are pulling. | 14 | 5 |
| Check For Negation | Determine if negated mentions of terms exist. In some instances, you may need to confirm a negative mention exists. In others, you may need to filter out terms that are negated. | 15 | 6 |
| Confirm Disease Das Checked | Make sure the patient has been in to see a doctor to be screened for a condition. This may also apply to labs, to ensure that a lab value was checked & came back normal. | 15 | 6 |
| Use Distinct Dates | When requiring a count of items, make sure they happen on multiple dates, possibly with some time interval between them. | 28 | 11 |
| Rule of N | More evidence is often required, especially when recurring codes gives a higher level of certainty that a condition exists and wasn't just a rule-out. | 39 | 14 |

## B. Data processing and feature extraction

For the application of ML algorithms on the classification of design patterns, we focused on 6 phenotype design pattern classes that have enough supporting sentences: "Confirm Disease Was Checked", "Rule of N", "Use Distinct Dates", Where Did It Happen", "Credentials of the Actor", and "Check For Negation". The description and number of sentence fragments for each design pattern are shown in Table 1. The total data consist of 131 fragments with 653 unique words.

We substituted all the numbers with "_number" and removed stop words. We experimented with features including TFIDF-weighted bag-of-words model, developing sites of the phenotype algorithms, phenotype (e.g., T2DM), the Unified Medical Language System (UMLS) concept unique identifiers (CUIs), and UMLS semantic types (STs). The site and phenotype features were included to account for the possible site- and phenotype associated biases. The CUIs and STs were obtained by running MetaMap [8] over the text fragments and were included to solve the word sparsity problem. We used the occurrence of each CUI or ST as the feature.

Observing that synonyms may occur frequently in narratives, we also experimented with word embedding that was trained on a large corpus of EHRs from the MIMIC III dataset using the word2vec toolkit [9] with the embedding dimension of 200. The feature matrix for the embedding model was created by the matrix multiplication of the bag-of-words matrix and the embedding matrix.

## C. Classification with Supporter Vector Machine SVM

Because each sentence may belong to multiple design pattern classes, we trained a one-vs-all classifier for each class. We split the train and test set at a 7 to 3 ratio. We used the linear kernel SVM algorithm as it has demonstrated good performance on sentence classification tasks as well as great generalizability. We scaled the input feature vector to its $l2$ norm to avoid scale invariant problem. We used the inverse of the class frequency to give higher weight to the underrepresented class. We used three-fold cross-validation to tune two hyper parameters (penalty parameter C [$10^{-6}, 10^3$] and penalty forms[$l1, l2$]) by optimizing the area under rectile eiver operating characteristic (ROC) curve. We applied the cross-validation-chosen best model on the test set and calculated f1 score for each class. We used micro-f1 and macro-f1 scores to evaluate the model performance across multiple classes. While macro-f1 is an arithmetic average of f1 scores across classes, micro-f1 takes account of class sizes and is essentially instance level average.

## D. Classification with convolutional neural network (CNN)

With the same train and test split, we also trained a convolutional neural network (CNN) classifier for each design pattern. Recently, deep learning methods have been successfully applied to text classification, and one such representative deep learning model is the convolutional neural network. Studies showed that CNN achieved impressive performance in general domain NLP [10]. In particular, several authors showed that CNN can perform comparably to state-of-the-art systems without heavy feature engineering on short clinical text classification tasks [11-13]. Inspired by such successes, we adapted a CNN model for design pattern classification. We used 20% of the training set as the validation set. We used the embedding vectors trained from MIMIC III with a range of embedding dimensions ([100, 200, 300, 400, 500, 600])[14]. Our CNN model takes the word embeddings of the sentence segments as input, performs convolution and max-pooling, and finally outputs the predicted design pattern annotation. We used a filter size of [3, 4, 5] and set the number of filters to be 50 given our small data size and short sentence length. We used 15 as the batch size and trained for 200 epochs. The dropout rate was 0.5 and the learning rate was 0.001. The reported f1 score was on the test set.

## III. RESULTS

### A. Visualizing relations among design patterns, phenotypes, and sites

We visualized relations among design patterns, phenotypes and sites using the site-phenotype network (Figure 1), pattern-phenotype network (Figure S1) and pattern-site network (Figure 2). From Figure 1, we see that the number of phenotypes corresponds to the phase when each site was added to the eMERGE network, of the 3 phases so far. Sites that started in phase I are Northwestern, Vanderbilt, Mayo, Group Health, and Marshfield. Geisinger, Mount Sinai, and Cincinnati Children's Hospital Medical Center (CCHMC). Children's Hospital of Philadelphia (CHOP) joined in phase II, and Columbia joined as an affiliate site. In phase III, Mount Sinai was no longer part of the network, and Columbia was an official site. We noted that the data set used represented phenotypes from phase I and II of eMERGE. From Figure 2, we observed that "Rule of N" and design patterns appear to be the most used patterns across sites, phenotypes, sentences in phenotyping algorithm. They are often used together in algorithms to specify the inclusion or exclusion events occurred on more than N number of distinct days. In addition, "Check for Negation", "Confirm Disease" and "Level of Evidence" are more highly represented in both Table 2 and Figure S1. We also noted that several sites and phenotypes are associated with more design patterns. We posit that this is in part an effect of the general disease category, which in turn is related to how and where the data are recorded in the EHR. For example, many of Marshfield's phenotypes are ophthalmological (see Figure 1), which was not as routinely or clearly recorded in a structured format. That would explain why more patterns are associated with Marshfield (see Figure 1). When checking the sentences that entail the design patterns, we found that aggregate "count" appeared to be the most common use of distinct dates and "confirm disease was checked" appeared to be commonly associated with lab or other measurements, which are used by clinicians to check for disease. For "Rule of N", the most common N is 2 (more than 60 mentions), while other Ns seem to have frequency exponentially decreased with N. Only two phenotypes from two sites (Cataract from Marshfield and Dementia from Group Health) used "Local Codes", which could be indicative of the fact that these phenotype algorithms were designed to be portable across multiple sites. As can be seen in the figures, although the adoption of design patterns is wide across eMERGE networks, the adoption is nevertheless heterogeneous at both the across-phenotype level and across-site level. For example, multiple phenotypes (e.g., MS, Heart Failure, Autism etc.) use only one design pattern, while certain phenotypes use many more design patterns (e.g., 10 for HDL). The same is true when inspecting pattern-site network (e.g., compare Marshfield with CCHMC).

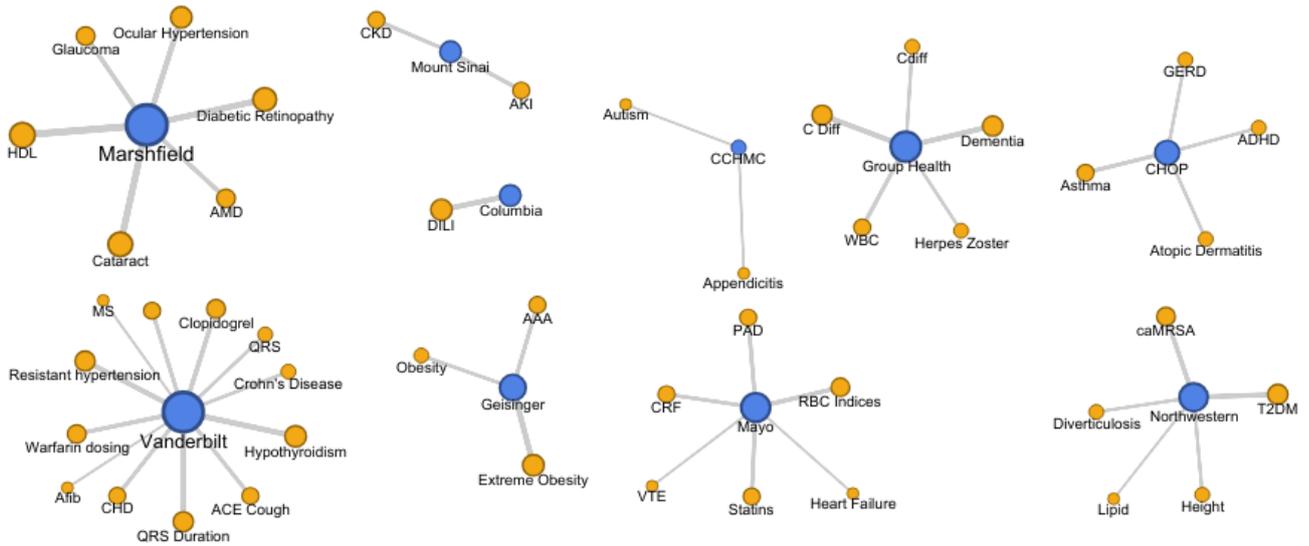

Figure 1. Network of sites and phenotypes. The size of the node is proportional to their times of occurrence in the dataset and the width of the edge corresponds to the number of connections between the associated nodes. CCMHC: Cincinnati Children's Hospital Medical Center.

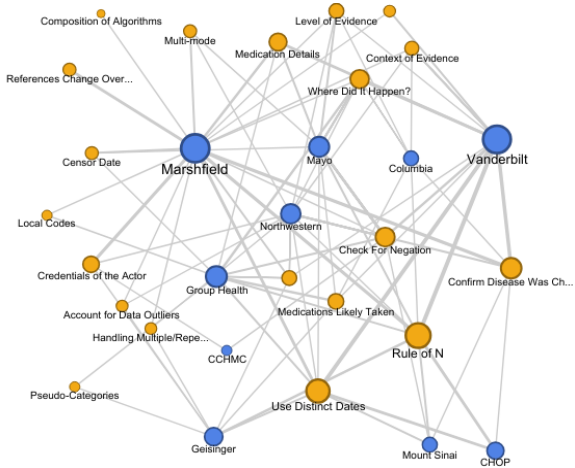

Figure 2. Network of design patterns and sites.

### B. Classfication

The maximum micro- f1 and macro-f1 for classification is 0.77 and 0.79 respectively using features of bag-of-words and STs (Table 2). Notably, each class achieved its best score with different models. For example, "Rule of N" has the best f1 with the baseline bag-of-words model, while adding the phenotype gives the perfect classification for "Use Distinct Dates" and "Credentials of the Actor". Embedding model did not outperform the bag-of-words model and the best performance was achieved with CUI as the additional feature of 0.59 macro-f1 and 0.55 micro-f1. We visualized the word embedding vectors through t-SNE clustering which showed that words of similar meanings were close to each other (Figure S2). We see that although some words clustered together in an intuitive way (e.g., days, months, years, weeks in the lower left corner), other words tend to form a mixed cloud where it may be difficult to see patterns, (e.g., in the middle of the cloud). This reflects the characteristics of clinical notes where word semantics are probably of high dimension On the other hand, the modestly high performance of bag-of-words models suggests that the design patterns may be signified by a collection of words themselves (often reflecting a topic in the context of topic modeling rather than nuanced semantics).

With the CNN model, the best performance was observed with the embedding dimension of 500 and 600, even though there is no consistent improvement with the increasing of the embedding dimension. Comparing the CNN model and embedding model that both are based on word vector representations, we observed that the CNN model always outperformed the embedding model for the "Credentials of the Actor" class.

Table 2. F1 scores for phenotyping design pattern classification. Included are results with bag-of-words/embedding model (with different additional features) and CNN model (with different word embedding dimensions, showing embedding dimension of 600). Abbreviations used: BOW: bag-of-words; CUI: UMLS concept unique identifiers; ST: UMLS semantic types; Pheno: targeted phenotype. Bold indicate best results. Bold indicates best performance in each column.

| Experiments | Disease | Rule of N | Dates | Credential | Where | Negation | Macro-f1 | Micro-f1 |
|---|---|---|---|---|---|---|---|---|
| **BOW** | 0.44 | **0.83** | 0.80 | 0.50 | 0.52 | 0.77 | 0.68 | 0.66 |
| **BOW + Site** | 0.00 | 0.73 | **1.00** | **1.00** | 0.77 | 0.72 | 0.71 | 0.76 |
| **BOW + Pheno** | 0.00 | 0.76 | 0.82 | 0.38 | **0.80** | 0.60 | 0.61 | 0.62 |
| **BOW + CUI** | 0.29 | 0.80 | 0.82 | 0.44 | 0.55 | 0.73 | 0.71 | 0.66 |
| **BOW + ST** | **0.50** | 0.74 | 0.82 | **1.00** | 0.71 | **0.83** | **0.79** | **0.77** |
| **Embedding** | 0.31 | 0.67 | 0.43 | 0.57 | 0.50 | 0.43 | 0.52 | 0.49 |
| **Embedding + Site** | 0.43 | 0.59 | 0.38 | 0.67 | 0.43 | 0.67 | 0.54 | 0.52 |
| **Embedding + Pheno** | 0.31 | 0.67 | 0.43 | 0.57 | 0.50 | 0.43 | 0.52 | 0.49 |
| **Embedding + CUI** | 0.25 | 0.79 | 0.30 | 0.73 | 0.71 | 0.44 | 0.59 | 0.55 |
| **Embedding + ST** | 0.00 | 0.67 | 0.39 | 0.73 | 0.62 | 0.59 | 0.52 | 0.52 |
| **CNN-600** | 0.29 | 0.48 | 0.57 | 0.89 | 0.44 | 0.60 | 0.60 | 0.54 |

## IV. DISCUSSION

We observed the heterogeneous popularity of design patterns in the phenotype algorithms. The most popular design pattern is the "Rule of N" with 53 supporting sentences. This is not surprising as meeting the quantitative criteria is essential for accurate and reproducible phenotyping. Interestingly, the number of sentences is not the most important factor in determining the algorithm performance. Notably, the bag-of-words model outperformed the embedding and CNN model. This is likely due to the fact that the sentence fragments extracted phenotype algorithm may feature repeated keywords that serve as a cue for suggesting design patterns (e.g. separate, apart for Use Distinct Dates; hospital, inpatient, outpatient for Where Did It Happen). Another potential explanation is that we have a relatively small dataset that cannot support training a CNN model that fully distinguishes the nuanced semantics to its full power. The potential improvements are attainable as more phenotypes are created within the eMERGE network and other collaborative. We found UMLS semantic types help to capture the most missing information among all additional features. This suggests mapping to UMLS semantic types can be helpful in characterizing the context of the phenotyping steps.

## V. CONCLUSION

Design patterns extracted from phenotyping algorithms reflect a wealth of experience and knowledge about the phenotyping algorithm development. We automated the design pattern annotation process in order to help with the rationalization and standardization of phenotype algorithm development amid the nuances and complexities of working with EHR data. We experimented with feature engineering for classification and assessed the efficacy of features including bag-of-words, word embeddings, as well as knowledge-based features. The systematic experiments and evaluations presented here are intended as a starting point for articulating and documenting automated and generalizable phenotyping design pattern identification. The overall performance demonstrates the practicability and challenges of automatic identification of design patterns and points to needs for more phenotype algorithms. We expect the broader biomedical informatics community to find the task and approach interesting and continue to investigate it at a larger scale.